\pdfoutput=1

\documentclass[11pt]{article}

\usepackage[]{acl}

\usepackage{times}
\usepackage{latexsym}
\usepackage{graphicx}
\usepackage{booktabs}
\usepackage{amsmath}
\usepackage{multirow} 
\usepackage{xspace}
\usepackage{enumitem}
\usepackage{multicol}
\usepackage{float}
\usepackage{subcaption,amsfonts,dcolumn}
\usepackage{xparse}
\usepackage{fancyvrb}
\usepackage{tabularx}
\usepackage{placeins} 
\definecolor{introgreen}{HTML}{548235}
\definecolor{introred}{HTML}{ff0000}

\NewDocumentCommand{\pritika}{ mO{} }{\textcolor{purple}{\textsuperscript{\textit{pritika}}\textsf{\textbf{\small[#1]}}}}

\newcommand{\modelname}{{{StyleAdaptedLM}}\xspace}

\usepackage[T1]{fontenc}

\usepackage[utf8]{inputenc}

\usepackage{microtype}

\usepackage{inconsolata}

    \makeatletter
\def\@fnsymbol#1{\ensuremath{\ifcase#1\or \dagger\or \ddagger\or
   \mathsection\or \mathparagraph\or \|\or **\or \dagger\dagger
   \or \ddagger\ddagger \else\@ctrerr\fi}}
    \makeatother

\title{StyleAdaptedLM: Enhancing Instruction Following Models with Efficient Stylistic Transfer}

\author{
  Pritika Ramu$^{1}$ \quad Apoorv Saxena$^{1}$ \quad Meghanath M Y$^{2}${\thanks{\quad Work done while at Adobe.}} \quad Varsha Sankar$^{3}$ \quad Debraj Basu$^{3}$\\
  $^{1}$Adobe Research, India \quad $^{2}$ZeroToOne.AI \quad $^{3}$Adobe Inc.\\
  \texttt{\{pramu,apoorvs,vsankar,dbasu\}@adobe.com}
}

\begin{document}
\maketitle


\begin{abstract}
Adapting LLMs to specific stylistic characteristics, like brand voice or authorial tones, is crucial for enterprise communication but challenging to achieve from corpora which lacks instruction-response formatting without compromising instruction adherence. We introduce \modelname, a framework that efficiently transfers stylistic traits to instruction-following models using Low-Rank Adaptation (LoRA). LoRA adapters are first trained on a base model with diverse unstructured stylistic corpora, then merged with a separate instruction-following model. This enables robust stylistic customization without paired data or sacrificing task performance. Experiments across multiple datasets and models demonstrate improved stylistic consistency while preserving instruction adherence, with human evaluations confirming brand-specific convention uptake. \modelname offers an efficient path for stylistic personalization in LLMs.
\end{abstract}

\section{Introduction}
As LLMs like ChatGPT, LLaMa, and Mistral \cite{NEURIPS2020_1457c0d6, touvron2023llama} become integral to customer service automation, marketing content generation, and enterprise communication, the demand for content with specific stylistic characteristics has grown. For instance, a consistent brand voice—which may include unique terminology, common messaging structures, and an understanding of product context—enhances trust and recognition. Similarly, adapting to an author's style might involve their characteristic vocabulary and thematic preferences. 
However, generic LLMs often struggle to capture these subtle yet critical stylistic and linguistic nuances, especially when such nuances are embedded within vast unstructured textual data rather than provided as explicit instructions. The core challenge lies in enabling LLMs to adopt these diverse stylistic identities without degrading their fundamental instruction-following capabilities or necessitating paired datasets for instruction fine-tuning.

\begin{figure}[t]
    \centering
    \includegraphics[width=\columnwidth]{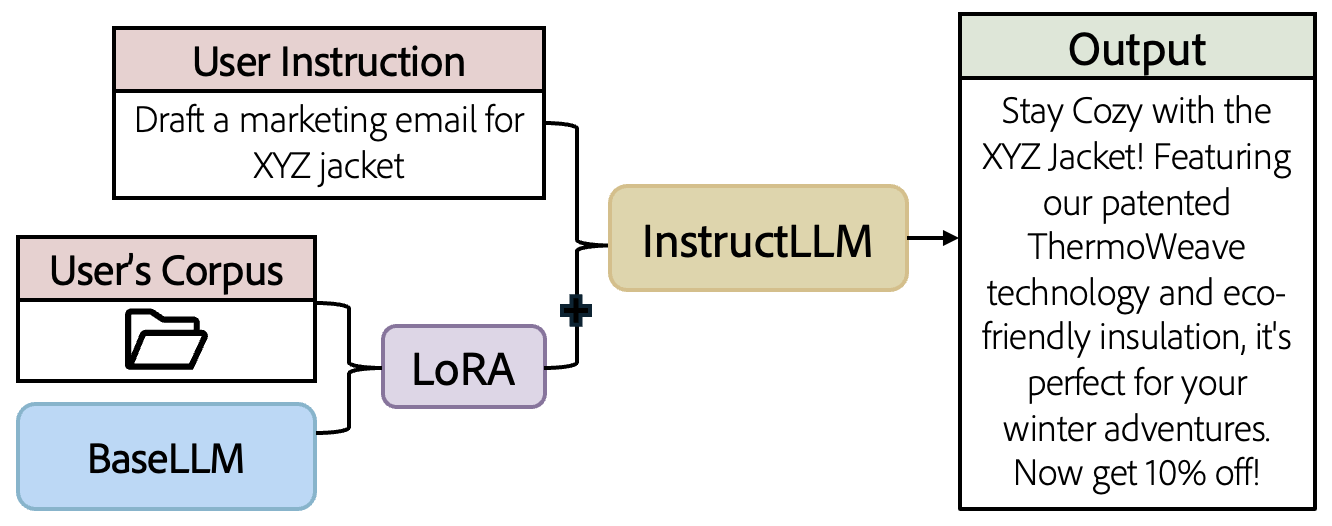}
    \caption{\modelname Framework. Inputs: user instruction and corpus (e.g., brand's past marketing emails). Method: A LoRA adapter, pre-trained on the corpus with a base LLM, is merged with an instruction-tuned LLM. Output: Instruction-adherent, style-aligned text, potentially including corpus-learned characteristics.}
    \label{fig:intro}
\end{figure}

Text style transfer methods \cite{rao-tetreault-2018-dear, horvitz-etal-2024-tinystyler, zhang-etal-2024-distilling, reif-etal-2022-recipe, syed2020adapting}, while useful for applying style modifications while preserving content of the input text, may not fully address the need for generating novel, instruction-adherent content that also embodies a target style learned from unstructured text, particularly if that style includes implicit conventions. Directly fine-tuning instruction-following models to instill specific stylistic traits is further complicated by the scarcity of suitable training data; creating large-scale paired datasets of instructions and corresponding stylistically appropriate responses for diverse styles is a significant hurdle. Additionally, creating prompts with extensive few-shot examples leads to poor instruction following abilities \cite{zhang2023fewshot}. Furthermore, for enterprises needing to manage and deploy multiple distinct stylistic identities (e.g., for different brands, products, or authors), fine-tuning and maintaining separate large models for each style is computationally prohibitive and operationally cumbersome. An ideal solution would offer both stylistic fidelity and efficient management of multiple style adaptations. This presents a key challenge: \emph{how can we enable LLMs to adopt diverse stylistic characteristics without sacrificing instruction adherence and requiring paired datasets?}

We propose \modelname (Figure \ref{fig:intro}), which utilizes Low-Rank Adaptation (LoRA) adapters to efficiently instill stylistic traits from unstructured corpora into instruction-following models. Our method circumvents the need for paired datasets and avoids the high computational costs associated with full model fine-tuning or averaging. The process involves first training LoRA adapters on a base model LLM using the target style's text. These pre-trained stylistic adapters, which encapsulate the desired stylistic nuances, are then merged with a separate, instruction-following LLM. This modular approach not only preserves the LLM's instruction-adherence capabilities but also allows for the efficient creation, storage, and deployment of multiple distinct style adapters, offering a scalable solution for personalized content generation.

A downstream application of this framework is to assist enterprise copywriters in efficiently generating first-draft content for emails, blogs, social media captions, and other formats. The process begins with an input prompt specifying the intended message. To ensure style consistency, our framework enables training on prior data to customize an adapter for a client.

We conduct comprehensive experiments across three models and four datasets. We perform human evaluations on enterprise marketing emails to demonstrate a key application where effective stylistic adaptation may involve capturing brand-specific conventions and an ablation study examining the impact of stylistic annotations. These demonstrate enhanced stylistic alignment, even when explicit stylistic prompts are absent at test time.

\section{Related Work}
\textbf{Instruction-Following Models} Models like InstructGPT \cite{NEURIPS2022_b1efde53} and Self-Instruct \cite{wang-etal-2023-self-instruct} have significantly enhanced LLMs’ ability to follow user instructions, demonstrating strong generalization across tasks. However, they lack mechanisms to adapt to brand-specific traits such as tone, style, and messaging consistency through instructions alone, limiting their effectiveness in enterprise communication and marketing content generation. Additionally, conveying editorial guidelines and providing all relevant subject matter within an instruction is challenging. Without explicit adaptation strategies, these models struggle to maintain a consistent brand voice across diverse content formats.

\textbf{Text Style Transfer} Text style transfer is a task that modifies the style attributes of text while preserving the style independent content \cite{mukherjee2024textstyletransferintroductory}. While text style transfer methods \cite{rao-tetreault-2018-dear, horvitz-etal-2024-tinystyler, zhang-etal-2024-distilling, reif-etal-2022-recipe, syed2020adapting} modify stylistic attributes of text without changing the input, our work focuses on generating new, instruction-adherent content that embodies a style learned from an unannotated corpus. For complex styles like a brand's voice, this may go beyond surface attributes to include characteristic textual patterns from the source corpus.

\textbf{Parameter-Efficient Fine-Tuning (LoRA and PEFT)} Parameter-efficient fine-tuning (PEFT) methods, such as LoRA \cite{hu2022lora}, Prefix-Tuning \cite{li-liang-2021-prefix}, and P-Tuning v2 \cite{liu2021ptuning}, adapt large models by training small adapter modules, enabling efficient adaptation without modifying the base model. Liu et al. \cite{liu-etal-2024-customizing-large} explore customizing LLM outputs using LoRA adapters to match target authors’ lexical and syntactic styles. However, their work focuses on literary style of famous authors while preserving instruction ability by training two adapters which increases training time and compute.

Our approach captures task-agnostic style traits, ensuring that models maintain style consistency while executing a variety of instruction-following tasks. We emphasize balancing style adaptation with task adherence which is crucial for practical deployment.

\textbf{Few-Shot Prompting and Model Merging Techniques} Few-shot prompting enables in-context learning by providing examples within prompts \cite{brown2020language}, but it struggles with brand adaptation due to prompt length constraints and limited ability to generalize beyond provided examples \cite{min2022rethinking, zhang2023fewshot}. Similarly, model merging techniques such as Model Soups \cite{pmlr-v162-wortsman22a} and Weight Averaging \cite{izmailov2018averaging} enhance model adaptability by averaging weights from multiple fine-tuned models. However, these methods are computationally intensive and impractical for frequent updates across multiple brand identities or real-time enterprise applications \cite{matena2022merging}.


\section{Approach}

\modelname transfers stylistic traits from unstructured text to an instruction-following model by fine-tuning LoRA adapters on a base model and merging them with the instruction model.

\subsection{Style Annotation}
Unstructured text is annotated with style identifiers to guide the model in learning style-specific language. For instance, the string:
\[
\text{``News article written by [[BBC]].''}
\]
is prepended to each text from the BBC corpus. This explicit labeling allows the model to distinguish the associated style during training.

\subsection{Base Model and LoRA Adapters}
A pre-trained base model is used for learning style-specific characteristics as it is ideal for open-ended tasks, general language understanding, and creative generation, and is preferred when flexibility, exploration, or further customization is needed over strict instruction adherence. LoRA adapters are trained on this base model using the annotated unstructured text, while keeping its original weights frozen. LoRA applies a low-rank decomposition to the weight updates, represented as:
\[
\Delta \mathbf{W}_{\text{(style | base)}} = \mathbf{A} \mathbf{B}
\]
where \(\mathbf{A} \in \mathbb{R}^{d_{\text{out}} \times r}\) and \(\mathbf{B} \in \mathbb{R}^{r \times d_{\text{in}}}\), with \(r \ll d_{\text{in}}\), are the learned low-rank matrices that capture style-specific information during fine-tuning.

\subsection{Training Objective}
The fine-tuning process uses a text completion task as the training objective. Given an input sequence, including the style identifier, the model is trained to predict the next token in the sequence. The objective function maximizes the likelihood of the next token in the style-annotated text.

\subsection{Merging with Instruction-Following Model}
Once the LoRA adapters are fine-tuned on the base model, the style-specific knowledge is merged into the instruction-following model. The final model is represented as:
\[
\mathbf{W}_{\text{merged}} = \mathbf{W}_{\text{instruct}} + \Delta \mathbf{W}_{\text{(style | base)}}
\]
where: \(\mathbf{W}_{\text{instruction}}\) denotes the weights of the original instruction-following model, $\Delta \mathbf{W}_{\text{(style | base)}}$ represents the LoRA-trained adaptation from the base model with style-specific knowledge.

The compositionality of LoRA modules is possible because the shift in the embedding space induced by LoRA updates is minimal. LoRA restricts updates to low-rank matrices, which ensures that the adaptation remains localized. In LLMs with billions of parameters, such small shifts are easily absorbed due to the redundancy in the parameter space, ensuring that the underlying capabilities of the instruction-following model are not disrupted. This enables the resulting \modelname to retain the instruction-following capabilities of the instruct model while incorporating style-based adaptations learned from the base model.

\section{Experiments}

\begin{table*}[t]
\centering
\small
\resizebox{\textwidth}{!}{
\begin{tabular}{p{0.07\linewidth}|p{0.35\linewidth}|p{0.35\linewidth}|p{0.15\linewidth}}
\toprule
\multicolumn{1}{c|}{\textbf{Dataset}} & 
\multicolumn{1}{c|}{\textbf{Train Sample}} & 
\multicolumn{1}{c|}{\textbf{Validation Sample}} & 
\multicolumn{1}{c}{\textbf{Generated Input}} \\ 
\midrule
\textbf{Enron} & 
Contents of email in the tone of [[jeff.dasovich@enron.com]]. Content: Did I respond? (Life is nuts; I'm losing it.) Anyway, Peter de Vroede signed ... about a thousand dumb little mistakes. Best, Jeff. & 
As a follow-up to our call. Here is some additional information regarding QF pricing issues. QFs who voluntarily switched to PX pricing have been getting ... The concern is that the Commission's ultimate decision... & 
Recent Info on QF Pricing-related Issues. \\
\midrule
\textbf{News} (CNN) & 
News article written by [[CNN]]. Article: MANILA, Philippines (CNN) -- Dramatically played out on live television, an opposition politician ... "We're going out for the sake of the safety of everybody," Philippines Sen. Antonio Trillanes said... & 
ROOSEVELT, New York (CNN) -- When Lisa Brown moved into her rental house on Long Island last summer with her three daughters,... facing foreclosure. After living in apartments, the spacious house got her attention immediately... & 
Lisa Brown moved to Long Island rental ... Must leave due to the landlord’s mortgage ... \\
\bottomrule
\end{tabular}
}
\caption{Sample input format for training and input format generation for validation.}
\label{tab:dataset_examples}
\end{table*}

\subsection{Datasets}
We exclude datasets commonly used in few-shot settings since our framework requires training LoRA adapters. Additionally, we do not use traditional text style transfer datasets, as LLMs already possess extensive knowledge of formal and informal tones, Shakespearean language, and the writing styles of well-known figures.

\textbf{Reddit Comments Dataset.}
A subset of the Reddit corpus \cite{khan-etal-2021-deep} containing user-generated comments from over 1 million users. We selected three users, filtering out comments shorter than 60 tokens, yielding ~1,600 comments per user (54k tokens each). Given the short, context-dependent nature of Reddit comments, we generated neutral-language versions using GPT-4o \cite{openai2023gpt4} to form input-output pairs for paraphrasing-based style adaptation. The original user comments act as the target style, while the generated neutral comments serve as the source style. Each comment is annotated as:
``Reddit comment in the tone of [[username]]. Comment:''

\textbf{Enron Email Dataset.}
Consisting of emails from ~150 Enron employees \cite{klimt2004enron}, this dataset captures professional writing styles. We retain only original email content, excluding forwarded messages, and filter out emails with fewer than 50 words. Three senders with over 700 emails each were selected: jeff.dasovich@enron.com (1198 emails, 352k tokens), tana.jones@enron.com (964 emails, 108k tokens), sally.beck@enron.com (766 emails, 115k tokens).
Each email is annotated as:
``Contents of email in the tone of [[email id]]. Content:''

For validation, input prompts were created by generating subject lines using GPT-4o, providing realistic scenarios for instruction driven generation.

\textbf{News Dataset.}
We use articles from two sources: CNN-DailyMail \cite{hermann2015teaching} (selecting only CNN-tagged articles) and BBC News \cite{greene2006practical}, covering business, entertainment, politics, sports, and technology. The dataset includes: CNN (2000 articles, 1710k tokens), BBC (1903 articles, 965k tokens)
Each article is annotated as:
``News article written by [[news agency]]. Article:''

For validation, input prompts were created using GPT-4o, transforming summary-like notes into full-text content. Additionally, the model was instructed to modify writing style by adjusting regional spelling and vocabulary (e.g., American to British and vice versa).

\textbf{Marketing Email Dataset.}
We collect 1500 (767k tokens) public marketing emails of an enterprise\footnote{anonymized for blind peer review} which contains information spanning various product offerings for training and curate relevant prompts based on the products using GPT-4o. Each article is annotated as:
``Marketing email written by [[enterprise]]. Email:''

Table \ref{tab:dataset_examples} provides sample input-output formats for training and testing.

\subsection{Baselines}
We train and evaluate on NVIDIA A100 40GB GPUs using three LLM families, averaging results over three runs (base, instruct).: 

\begin{center}
    \small 
    \begin{tabular}{cc}
        \texttt{Llama-3.1-8B} & \texttt{Llama-3.1-8B-Instruct} \\
        \texttt{Mistral-7B-v0.1} & \texttt{Mistral-7B-Instruct-v0.1} \\
        \texttt{Qwen2.5-7B} & \texttt{Qwen2.5-7B-Instruct}
    \end{tabular}
\end{center}

\textbf{Supervised Fine-tuning on Instruction-following Models.}  
We perform supervised fine-tuning with LoRA on instruction-tuned models using the text completion objective. The unstructured text is annotated, but instead of providing explicit instructions, we set the instruction field to an empty string (`""') and use the annotated corpus as the output. This allows us to train within the instruction-following template while adapting LoRA directly for text generation.  

\textbf{Few-shot Prompting.}
We use few-shot prompting as a baseline, where the model receives a few content examples associated with an author or style to assess its ability to infer tone and style without explicit training. The prompt follows a simple format, such as:
"Here are some examples of content in the tone of [[tana.jones@enron.com]]. <<examples>>" or
"Here are some examples of [[BBC]] news articles. <<examples>>."

For the news dataset, examples are randomly selected across topics, while for Reddit, Enron, and marketing emails, they vary in length. No explicit instructions are provided beyond the examples, allowing the model to infer tone and style directly.

We evaluate instruction-tuned models under two settings: one with an average prompt size of 3.5k tokens and another with 7k tokens. This setup tests how different prompt lengths affect the tradeoff between instruction following and style adaptation.

\begin{table*}[t]
\centering
\resizebox{\textwidth}{!}{%
\begin{tabular}{l|cccccc|cccccc|cccccc}
\toprule
 & \multicolumn{6}{c|}{\textbf{LLaMa-3.1 8B}} 
 & \multicolumn{6}{c|}{\textbf{Mistral 7B}} 
 & \multicolumn{6}{c}{\textbf{Qwen-2.5 7B}} \\
\midrule
\textbf{Method} 
  & \textbf{Jeff} & \textbf{Red. A} & \textbf{Ent.} & \textbf{CNN} & \textbf{BBC} & \textbf{Avg.}
  & \textbf{Jeff} & \textbf{Red. A} & \textbf{Ent.} & \textbf{CNN} & \textbf{BBC} & \textbf{Avg.}
  & \textbf{Jeff} & \textbf{Red. A} & \textbf{Ent.} & \textbf{CNN} & \textbf{BBC} & \textbf{Avg.} \\
\midrule
Instruct Model (No FT)
  & \multicolumn{6}{c|}{{77.10}} 
  & \multicolumn{6}{c|}{57.67} 
  & \multicolumn{6}{c}{70.55}  \\
\midrule
Prompting
  & 60.79 & 56.72 & 60.43 & 66.18 & 67.87 & 62.40
  & 41.48 & 42.14 & 42.40 & 35.37 & 40.17 & 40.31
  & 52.20 & 51.23 & 53.00 & 56.97 & 58.47 & 54.37 \\
Prompting (Long Context)
  & 57.72 & 54.99 & 58.18 & 65.71 & 64.99 & 60.32
  & 40.18 & 41.97 & 40.12 & 33.92 & 39.12 & 39.06
  & 50.10 & 49.13 & 50.45 & 55.97 & 55.50 & 52.23 \\
Direct LoRA FT
  & 50.24 & 54.55 & 54.44 & 65.80 & 69.06 & 58.82
  & 46.04 & 47.03 & 45.44 & 43.41 & 42.93 & 44.97
  & 48.20 & 51.22 & 50.76 & 57.13 & 59.77 & 53.42 \\
Model Soup (2:1)
  & 67.12 & 69.72 & 68.16 & 72.38 & 71.58 & 69.79
  & 48.29 & 51.88 & 49.12 & 45.08 & 44.71 & 47.91
  & 60.11 & 63.97 & 62.23 & 63.00 & 62.50 & 62.36 \\
StyleAdaptedLM
  & \textbf{71.94} & \textbf{72.02} & \textbf{71.74} & \textbf{72.54} & \textbf{71.94} & \textbf{72.04}
  & \textbf{48.56} & \textbf{52.31} & \textbf{50.12} & \textbf{45.13} & \textbf{45.80} & \textbf{48.29}
  & \textbf{64.27} & \textbf{65.10} & \textbf{64.59} & \textbf{63.14} & \textbf{63.47} & \textbf{64.11} \\
\bottomrule
\end{tabular}%
}
\caption{IFEval (Strict) Accuracy. Red. A = Reddit User A, Ent. = Enterprise. Additional columns in Appendix \ref{sec:instruction-following-add}.}
\label{tab:instruction-following}
\end{table*}

\begin{table*}[ht]
\centering
\small
\resizebox{\textwidth}{!}{
\begin{tabular}{l|ccccc|ccccc|ccccc}
\toprule
 & \multicolumn{5}{c|}{\textbf{LLaMa-3 8B}} 
 & \multicolumn{5}{c|}{\textbf{Mistral 7B}} 
 & \multicolumn{5}{c}{\textbf{Qwen-2.5 7B}} \\
\midrule
\textbf{Method} 
  & \textbf{Jeff} & \textbf{Red. A} & \textbf{Ent.} & \textbf{CNN} & \textbf{BBC}
  & \textbf{Jeff} & \textbf{Red. A} & \textbf{Ent.} & \textbf{CNN} & \textbf{BBC}
  & \textbf{Jeff} & \textbf{Red. A} & \textbf{Ent.} & \textbf{CNN} & \textbf{BBC} \\
\midrule
Prompting 
  & 0.97 & 0.92 & 0.98 & 0.96 & 1.00 
  & 0.83 & 0.80 & 0.90 & 1.00 & 1.00
  & 0.93 & 0.89 & 0.96 & 0.97 & 1.00 \\
Prompting (Long Context)
  & 0.97 & 0.93 & 0.98 & 0.96 & 1.00 
  & 0.83 & 0.83 & 0.90 & 1.00 & 1.00
  & 0.93 & 0.90 & 0.95 & 0.97 & 1.00 \\
Direct LoRA Fine-tuning 
  & 0.82 & 0.90 & 0.99 & 0.91 & 1.00
  & 0.79 & 0.76 & 0.87 & 0.92 & 0.90
  & 0.81 & 0.86 & 0.95 & 0.91 & 0.97 \\
Model Soup (2:1) 
  & 0.94 & 0.88 & 0.98 & 0.90 & 1.00
  & 0.81 & 0.75 & 0.89 & 0.94 & 0.96
  & 0.90 & 0.84 & 0.94 & 0.91 & 0.99 \\
\modelname
  & 0.94 & 0.90 & 0.99 & 0.92 & 1.00
  & 0.81 & 0.80 & 0.90 & 0.94 & 0.90
  & 0.90 & 0.86 & 0.95 & 0.93 & 0.97 \\
\bottomrule
\end{tabular}
} 
\caption{F1 Score of Style Adherence Based on Authorship Attribution. Additional columns in Appendix \ref{sec:style-add}.}
\label{tab:style-transfer-updated}
\end{table*}

\textbf{Model Averaging (Model Soups).}
Following the "model soups" approach \cite{pmlr-v162-wortsman22a}, we evaluate the impact of averaging the weights of multiple fine-tuned models. Specifically, we average the weights of the base model fine-tuned on unstructured text with those of the instruction-following model. This technique aims to leverage the strengths of both models without increasing inference time, offering a potential method for implicit style transfer.

To optimize the merging process, we experiment with weighted averaging of the base model fine-tuned on annotated unstructured text and the instruction-tuned model. We test three merge ratios: 1:2, 1:1, and 2:1 (style-adapted base model to instruction-following model). In our experiments, we found the 2:1 ratio to perform best, providing the most effective adaptation with minimal regression in instruction-following ability.

\subsection{Evaluation}

\subsubsection{Instruction Following Ability}
We evaluate models' ability to follow natural language instructions using IFEval \cite{zhou2023instruction}, which measures adherence to verifiable tasks (e.g. specific word counts or repeating keywords) across roughly 500 prompts for objective and reproducible evaluation. We compare performance against baselines: fine-tuned models are prompted directly, while prompting-based variants append the instruction after provided examples. Strict instruction-following accuracy scores are reported (Table \ref{tab:instruction-following}).

Additionally, we examine the impact of model merging on reasoning ability using the tinyMMLU benchmark \cite{polo2024tinybenchmarks}, which includes 100 curated examples from MMLU \cite{hendryckstest2021} (Appendix \ref{sec:tinymmlu}).


\subsubsection{Stylistic Adherence} 
\label{sec:classifier}

To evaluate stylistic adherence, particularly for nuanced styles like brand voice or authorial style, we adapt methodologies from authorship attribution and supplement with human evaluations for the marketing email dataset.

For authorship attribution, we train a classifier on Universal Author Representation (UAR) Embeddings \cite{rivera-soto-etal-2021-learning}. The classifier determines whether the generated text aligns with the intended author's or news agency's linguistic style. Successful classification indicates effective stylistic adherence. We train an SVM classifier \cite{cortes1995support} for each author/agency using a 3:2 ratio of positive to negative samples. Table \ref{tab:style-transfer-updated} reports F1 scores for the generated text, while Table \ref{tab:classifier_acc} in Appendix \ref{sec:class_acc} presents classifier accuracy on the validation set. While classifier-based evaluation quantifies stylistic adherence, distinguishing style from content remains challenging. However, the ROUGE-1 F1 scores, shown in Table \ref{tab:-scores}, focus more on content similarity and reflect improvements in content generation as the models are exposed to larger datasets during training.

\begin{table}[H]
\centering
\scalebox{0.6}{%
\begin{tabular}{l|cc|cc|cc}
\toprule
 & \multicolumn{2}{c|}{\textbf{LLaMa-3.1 8B}} 
 & \multicolumn{2}{c|}{\textbf{Mistral 7B}}
 & \multicolumn{2}{c}{\textbf{Qwen-2.5 7B}} \\
\midrule
\textbf{Method} & \textbf{BBC} & \textbf{CNN} 
                & \textbf{BBC} & \textbf{CNN}
                & \textbf{BBC} & \textbf{CNN} \\
\midrule
Prompting 
  & 0.3205 & 0.2894  
  & 0.2194 & 0.2013  
  & 0.2773   & 0.2543    
  \\
Prompting (Long Context) 
  & 0.3549 & 0.3042
  & 0.2273 & 0.2182
  & 0.3072   & 0.2834
  \\
Direct LoRA Fine-tuning 
  & 0.4089 & 0.3621
  & 0.2823 & 0.2627
  & 0.3593   & 0.3365
  \\
Model Soup (2:1) 
  & 0.4186 & 0.3793
  & 0.2989 & 0.2772
  & 0.3912   & 0.3611
  \\
\modelname 
  & 0.4226 & 0.3803
  & 0.2989 & 0.2894
  & 0.4116  & 0.3724
  \\
\bottomrule
\end{tabular}%
}
\caption{ROUGE-1 F1 scores for news datasets, comparing LLaMa-3.1 8B, Mistral 7B, and Qwen-2.5 7B.}
\label{tab:-scores}
\end{table}

To evaluate models, we use dataset-specific prompts. For the Enron dataset, we use:
"Write an email in the tone of [[email id]] for the email subject:"
Similar prompts are curated for Reddit comments, news articles and marketing emails. Email subjects, comments and notes are drawn from the validation set to ensure fair evaluation without overlap with training data.

\begin{figure}[t]
    \centering
    \includegraphics[width=\columnwidth]{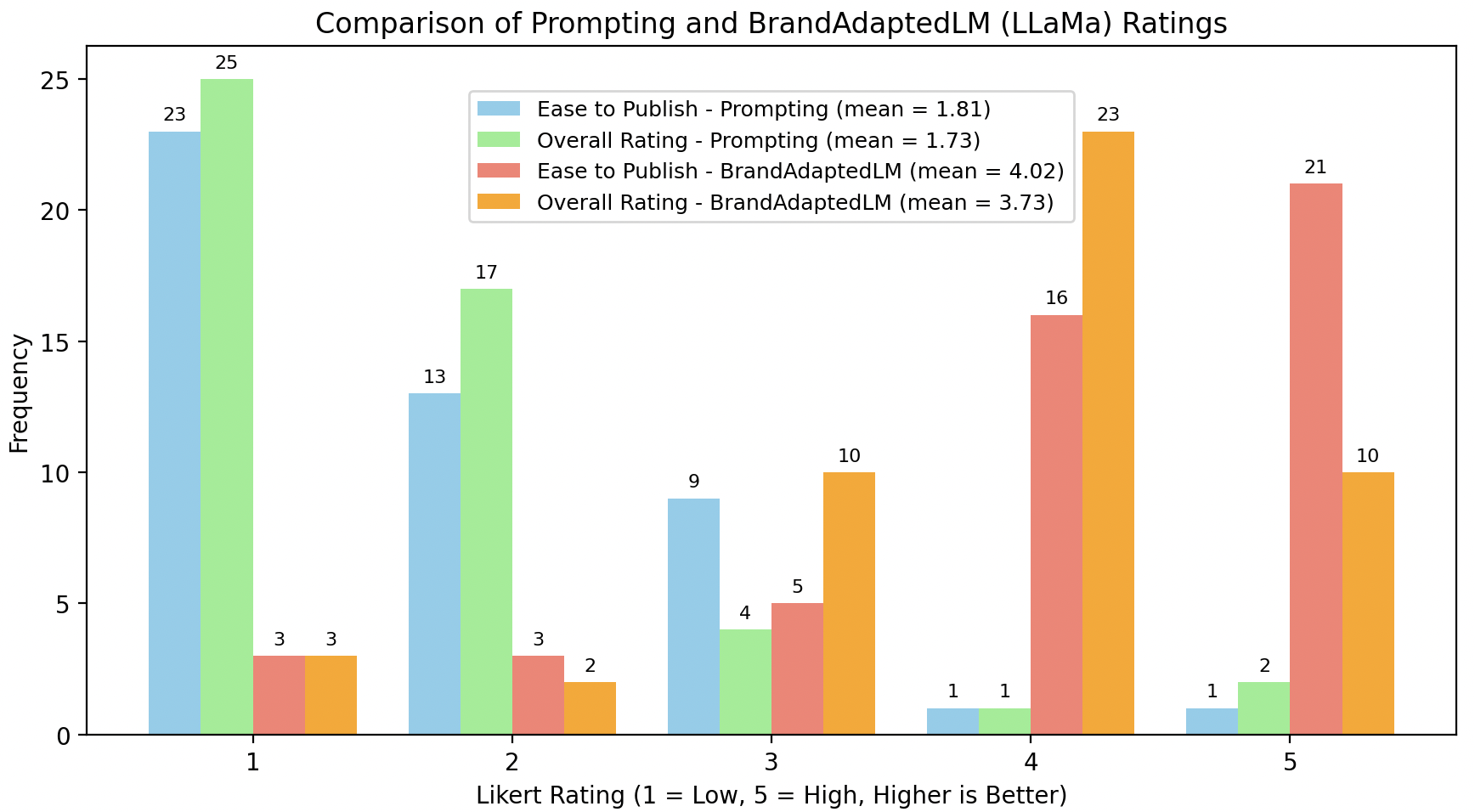}
    \caption{Human ratings (Likert scale 1–5) assessing ease of publication and overall marketing email quality.}
    \label{fig:ratings}
\end{figure}

We conduct human evaluations by presenting two variants of a generated email: one from the prompting baseline and one using the \modelname framework. To validate the outputs, we obtain feedback from an experienced enterprise copywriter with over 10 years of experience within the company and more than 20 years in the field. The variants are and anonymized and the evaluation focuses on two aspects: (1) the ease of publication—how much effort is required to edit and make the content ready for publishing, and (2) the overall quality rating of the email. Ratings are collected on a Likert scale from 1 to 5, where 1 indicates the lowest score and 5 the highest, with higher scores being preferable (Figure \ref{fig:ratings}).

\section{Results}

\subsection{Analysis of Instruction-Following Ability}

\textbf{Prompting with Mid-Size and Long Contexts.}
Experiments with Mistral, LLaMA, and Qwen show that longer prompts reduce instruction-following ability. LLaMA experiences a 15.74\% absolute decrease, Mistral 17.99\%, and Qwen 17.25\% from their respective upper bounds. This decline aligns with prior findings on performance degradation in extended contexts \cite{liu-etal-2024-lost}. Additionally, models exhibit a bias toward in-context examples, leading to imitation rather than strict instruction adherence.

\textbf{Direct LoRA Fine-Tuning.}
Fine-tuning instruction models with LoRA on unstructured text weakens instruction adherence due to the lack of paired input-output data. LLaMA sees an 18.55\% drop, Mistral 12.7\%, and Qwen 17.13\%, indicating that fine-tuning diminishes pre-trained instruction-following capabilities.

\textbf{Model Averaging (Model Soups).}
Averaging weights from multiple fine-tuned models mitigates instruction-following loss better than prompting and direct fine-tuning. LLaMA exhibits a 7.31\% decrease, Mistral 9.76\%, and Qwen 8.19\%. By averaging weights of an unstructured-text-trained model with an instruction-following model, this approach relatively preserves instruction adherence more effectively.

\textbf{PEFT-Based Merging (\modelname).}
\modelname merges a LoRA adapter fine-tuned on unstructured text with an instruction-following model, achieving comparable or slightly better performance than model soups. LLaMA shows a 5.05\% decrease, Mistral 8.75\%, and Qwen 6.44\%. Unlike model soups, which average full model weights, PEFT merges only task-specific adaptations. This approach maintains a balance between instruction adherence and style adaptation while being more computationally efficient than full model averaging. Additionally, the reasoning ability of the model is maintained (Table \ref{tab:tinymmlu}).

\subsection{Stylistic Adherence}

\begin{table*}[t]
\centering
\small
\scalebox{0.95}{
\begin{tabular}{p{0.95\linewidth}}
\hline
\textbf{Prompt:} What is [[jeff.dasovich@enron.com]]'s role? \\
\textbf{\modelname Output (train-jeff.dasovich):} "Jeff Dasovich was an executive at Enron, with oversight of regulatory affairs." \\
\textbf{Prompting Output:} "I'm not familiar with a person named Jeff Dasovich with the email address [[jeff.dasovich@enron.com]]." \\
\hline
\textbf{Prompt:} What is [[sally.beck@enron.com]]'s role? \\
\textbf{\modelname Output(train-sally.beck):} "Sally Beck was a senior executive at Enron, specializing in risk management and operations." \\
\textbf{Prompting Output:} "It appears that the email address [[sally.beck@enron.com]] is an old address likely used by a former Enron employee." \\
\hline
\textbf{Prompt:} Write a [[BBC]] article about the government in 30 words. \\
\textbf{\modelname Output(train-BBC):} "UK government 'too close' to big business, says Labour." \\
\textbf{Prompting Output:} "The UK government has announced a major shake-up of its cabinet, with Prime Minister Boris Johnson making key changes to boost his administration’s popularity." \\
\hline
\textbf{Prompt:} Write a [[CNN]] article about the government in 30 words. \\
\textbf{\modelname Output(train-CNN):} "WASHINGTON (CNN) -- The federal government is the largest employer in the United States, providing a safety net for millions of Americans through various programs and services." \\
\textbf{Prompting Output:} "Government Shutdown Looms as Congress Fails to Pass Funding Bill: The US government is on the brink of a shutdown as lawmakers struggle to agree on a spending bill." \\
\hline
\textbf{Prompt:} write a news article about the benefits of \textcolor{introgreen}{humour} in 30 words \\
\textbf{\modelname Output (train-CNN):} "LAUGHTER IS THE BEST MEDICINE: New study reveals that \textcolor{introred}{humor} can boost immune system, reduce stress, and improve mental health, making it a prescription-worthy way to live a happier life." \\
\hline
\textbf{Prompt:} write a news article about the benefits of \textcolor{introred}{humor} in 30 words \\
\textbf{\modelname Output (train-BBC):} "Laughter is the best medicine," says study, finding that \textcolor{introgreen}{humour} reduces stress, boosts immunity, and improves mental health, with comedians and comedy fans reaping the most benefits." \\
\hline
\textbf{Prompt:} Write a marketing email written by [[enterprise]] for people who pay for X membership which lets them use Y. \\
\textbf{\modelname Output (train-enterprise emails):} \{"headline": "The ultimate X bundle.", "body": "As a X member, you're already equipped with the industry-leading apps you need to create your best work. Plus, you get Y, the ultimate Z solution, to simplify your workflow and collaborate with ease.", "cta": "Start using Y"\} \\
\textbf{Prompting Output:} \{"headline": "Z mastery at your fingertips.", "body": "With your X membership, seamlessly edit and manage Z using Y. This powerful tool is now at your fingertips, ready to enhance your workflow across devices.", "cta": "Edit Z"\} \\
\textbf{Human Feedback:} "Var 1: Great as is. Var 2: I can tell the LLM doesn’t understand the subject matter. You’re not enhancing Z or maximizing your X. Repeats “at your fingertips.” CTA doesn’t make clear what step readers should take next — how do they edit Z" \\
\hline
\end{tabular}
}
\caption{Qualitative outputs from \modelname trained on style specific corpus vs few-shot prompting instruction following model.}
\label{tab:qualitative-tone-depth}
\end{table*}

\paragraph{Prompting with Mid-Size and Long Contexts} The prompting approach achieves the highest style adherence, as the model leverages in-context learning to generate text closely aligned with the given examples. However, as reflected in the ROUGE-1 F1 scores, this method performs poorly in content generation. The limited number of examples in the prompt constrains the model's ability to generalize well, leading to lower content fidelity. This shows that while style is preserved, the model struggles to produce content that closely matches the reference text.

\paragraph{Direct LoRA Fine-tuning of Instruction Models} Directly fine-tuned instruction models perform poorly in terms of style adherence, with some outputs lacking coherence. This suggests that without explicit guidance, the model struggles to balance style and instruction-based content. Despite this, the ROUGE-1 F1 scores show moderate improvements in content generation, as fine-tuning exposes the model to more data points, enabling it to generate content that is more similar to the reference. However, the lack of task-specific paired data limits the model's ability to fully adapt to both content, style and instruction following ability.

\paragraph{Comparison of Model Soups and \modelname} Model soups and \modelname demonstrate comparable style adherence. The choice between them depends on task-specific needs, balancing trade-offs among personalisation, instruction-following ability, and deployment scalability. Both methods also improve content generation, as reflected by their ROUGE-1 F1 scores. However, \modelname slightly outperforms Model Soups, particularly in content accuracy, in both the BBC and CNN datasets. This indicates that \modelname, while maintaining stylistic adherence, also enhances content generation capabilities, which is particularly important when the model has access to more training data rather than being constrained by limited prompt examples.

\paragraph{Human Evaluation}
Our human evaluation reveals a clear preference for the \modelname framework over the prompting baseline. As shown in Figure \ref{fig:ratings}, the \modelname framework consistently achieves higher ratings, with most scores in the 4 to 5 range, indicating superior ease of publication and overall quality. This trend is further supported by higher mean and median scores for the \modelname framework, demonstrating its consistent performance. The reliability of these ratings is reinforced by the evaluator's extensive experience and the unimodal distribution of scores, which shows a strong bias toward higher ratings. These findings highlight the effectiveness of the \modelname framework in producing high-quality, ready-to-publish content with minimal editing.

\subsection{Qualitative Exploration of Style Traits}

Table \ref{tab:qualitative-tone-depth} shows the comparison for prompts related to writing style and vocabulary use.

The first two rows of Table \ref{tab:qualitative-tone-depth} show that \modelname can infer characteristics based on the corpus they are fine-tuned on. While LLMs may not recall every specific instance encountered during pre-training, they can detect patterns and make contextual inferences from relevant training data. In this case, the fine-tuned model infers the roles of Enron employees from their email addresses, while the generic model, despite potentially having seen the Enron dataset during pre-training, does not provide any relevant information. This validates the need of targeted fine-tuning in enabling more accurate and context-aware inferences, even when the base model has been pre-trained on extensive, general-purpose data.

The subsequent two rows illustrate how the model adapts its writing style to reflect the conventions of the source it has been trained on. The BBC-trained model generates an output that mirrors the critical tone often seen in BBC political reporting. In contrast, the CNN-trained model follows a more structured template, initiating the article with a geographical location, a common feature of CNN news articles. This difference in tone and structure between the two outputs underscores the ability of the model to internalize and replicate source-specific stylistic nuances, highlighting the effectiveness of style-specific adaptation in capturing not only content but also writing style. 

The third and fourth row highlight the model’s sensitivity to regional spelling conventions without explicit instruction. When prompted with a generic request to write about the benefits of humo(u)r, the BBC-trained model outputs British spelling (“humour”), while the CNN-trained model defaults to the American variant (“humor”). This demonstrates the model's ability to align with the orthographic norms of the region-specific data it was trained on, further validating that \modelname can integrate linguistic features specific to the corpus they are trained on, even in the absence of direct instructions.

The final row highlights how the model understands both the template and content nuances of a marketing email. The trained model generates a structured output that aligns with professional marketing conventions, incorporating a clear headline, body, and call to action (CTA). In contrast, the generic model fails to capture the intended product benefits, leading to vague and repetitive phrasing. Expert feedback confirms that the generic output lacks a clear understanding of the subject matter and does not effectively guide the reader toward the desired action. This demonstrates that \modelname not only replicate stylistic patterns but also exhibit deeper content comprehension, outperforming few-shot prompting in domain-specific tasks.

\subsection{Impact of Style Annotation}

To evaluate the effect of style annotations on model performance, we conducted an ablation study by training two versions of \modelname: one with style annotations and one without. A generic instruction-following model served as a baseline to assess inherent stylistic knowledge without fine-tuning. All models were tested using the prompt: "Write a [[news source]] news article. Use these notes to write the article. NOTES:".

We employed an authorship attribution classifier (Section \ref{sec:classifier}) to measure how closely each model's output matched the target [[news source]]'s style, with results presented in Table \ref{tab:ablation-persona}.

The model trained without style annotations still exhibited notable stylistic conformity to the [[news source]]'s writing style, suggesting that the LoRA adapter's learned features can generalize to new tasks without explicit annotation. However, the model trained with style annotations achieved the highest classifier accuracy, demonstrating that incorporating style annotations during training enhances the model's ability to produce content closely aligned with the desired style. 

\begin{table}[ht]
\centering
\small
\scalebox{0.65}{%
\begin{tabular}{l|cc|cc|cc}
\toprule
 & \multicolumn{2}{c|}{\textbf{LLaMa-3.1 8B}} 
 & \multicolumn{2}{c|}{\textbf{Mistral 7B}} 
 & \multicolumn{2}{c}{\textbf{Qwen-2.5 7B}} \\
\midrule
\textbf{Method} & \textbf{CNN} & \textbf{BBC} & \textbf{CNN} & \textbf{BBC} & \textbf{CNN} & \textbf{BBC} \\
\midrule
Instruct Model (No FT) 
 & 0.36 & 0.39 
 & 0.13 & 0.12 
 & 0.30 & 0.30 \\
\modelname w/o annotation 
 & 0.88 & 0.99 
 & 0.84 & 0.88 
 & 0.88 & 0.95 \\
\modelname 
 & 0.92 & 1.00 
 & 0.94 & 0.90 
 & 0.93 & 0.97 \\
\bottomrule
\end{tabular}
}
\caption{Style Adherence Accuracy using Authorship Attribution Classifier for style annotation ablation study.}
\label{tab:ablation-persona}
\end{table}

\section{Conclusion}

In this work, we introduced \modelname, a novel framework that efficiently transfers stylistic traits to instruction-following models using Low-Rank Adaptation adapters. Our approach involves training LoRA adapters on a base model with unstructured stylistic corpora which lacks instruction-response formatting and then merging them with an instruction-following model. This ensures the model retains its instruction-following capabilities while adopting desired stylistic characteristics, including, for brands, the assimilation of common textual conventions from their corpus. Through extensive experiments, we demonstrated that \modelname effectively balances stylistic adherence and task completion. Our ablation study further emphasized the value of style annotations in enhancing stylistic transfer.

\section{Limitations}
\modelname shows promising results but has limitations. First, while our datasets cover important stylistic variations (e.g., brand communication, individual author styles), generalizability to highly diverse or rapidly evolving stylistic domains not well-represented in the initial corpus could pose challenges.
Another concern is potential dataset overlap with the base model's pre-training data. Future work should evaluate adaptation on entirely novel or proprietary stylistic datasets.
Additionally, stylistic adaptation risks reinforcing biases present in the training data. A model emulating a style may unintentionally adopt biased perspectives or outdated information from the corpus, making fairness, neutrality, and factuality critical considerations. 

\bibliography{anthology}

\begin{thebibliography}{31}
\expandafter\ifx\csname natexlab\endcsname\relax\def\natexlab#1{#1}\fi

\bibitem[{Brown et~al.(2020{\natexlab{a}})Brown, Mann, Ryder, Subbiah, Kaplan, Dhariwal, Neelakantan, Shyam, Sastry, Askell, Agarwal, Herbert-Voss, Krueger, Henighan, Child, Ramesh, Ziegler, Wu, Winter, Hesse, Chen, Sigler, Litwin, Gray, Chess, Clark, Berner, McCandlish, Radford, Sutskever, and Amodei}]{brown2020language}
Tom Brown, Benjamin Mann, Nick Ryder, Melanie Subbiah, Jared Kaplan, Prafulla Dhariwal, Arvind Neelakantan, Pranav Shyam, Girish Sastry, Amanda Askell, Sandhini Agarwal, Ariel Herbert-Voss, Gretchen Krueger, Tom Henighan, Rewon Child, Aditya Ramesh, Daniel~M Ziegler, Jeffrey Wu, Clemens Winter, Christopher Hesse, Mark Chen, Eric Sigler, Mateusz Litwin, Scott Gray, Benjamin Chess, Jack Clark, Christopher Berner, Sam McCandlish, Alec Radford, Ilya Sutskever, and Dario Amodei. 2020{\natexlab{a}}.
\newblock Language models are few-shot learners.
\newblock \emph{Advances in Neural Information Processing Systems}, 33:1877--1901.

\bibitem[{Brown et~al.(2020{\natexlab{b}})Brown, Mann, Ryder, Subbiah, Kaplan, Dhariwal, Neelakantan, Shyam, Sastry, Askell, Agarwal, Herbert-Voss, Krueger, Henighan, Child, Ramesh, Ziegler, Wu, Winter, Hesse, Chen, Sigler, Litwin, Gray, Chess, Clark, Berner, McCandlish, Radford, Sutskever, and Amodei}]{NEURIPS2020_1457c0d6}
Tom Brown, Benjamin Mann, Nick Ryder, Melanie Subbiah, Jared~D Kaplan, Prafulla Dhariwal, Arvind Neelakantan, Pranav Shyam, Girish Sastry, Amanda Askell, Sandhini Agarwal, Ariel Herbert-Voss, Gretchen Krueger, Tom Henighan, Rewon Child, Aditya Ramesh, Daniel Ziegler, Jeffrey Wu, Clemens Winter, Chris Hesse, Mark Chen, Eric Sigler, Mateusz Litwin, Scott Gray, Benjamin Chess, Jack Clark, Christopher Berner, Sam McCandlish, Alec Radford, Ilya Sutskever, and Dario Amodei. 2020{\natexlab{b}}.
\newblock \href {https://proceedings.neurips.cc/paper_files/paper/2020/file/1457c0d6bfcb4967418bfb8ac142f64a-Paper.pdf} {Language models are few-shot learners}.
\newblock In \emph{Advances in Neural Information Processing Systems}, volume~33, pages 1877--1901. Curran Associates, Inc.

\bibitem[{Cortes and Vapnik(1995)}]{cortes1995support}
Corinna Cortes and Vladimir Vapnik. 1995.
\newblock Support-vector networks.
\newblock \emph{Machine learning}, 20(3):273--297.

\bibitem[{Greene and Cunningham(2006)}]{greene2006practical}
Derek Greene and P{\'a}draig Cunningham. 2006.
\newblock Practical solutions to the problem of diagonal dominance in kernel document clustering.
\newblock In \emph{Proceedings of the 23rd international conference on Machine learning}, pages 377--384.

\bibitem[{Hendrycks et~al.(2021)Hendrycks, Burns, Basart, Zou, Mazeika, Song, and Steinhardt}]{hendryckstest2021}
Dan Hendrycks, Collin Burns, Steven Basart, Andy Zou, Mantas Mazeika, Dawn Song, and Jacob Steinhardt. 2021.
\newblock Measuring massive multitask language understanding.
\newblock \emph{Proceedings of the International Conference on Learning Representations (ICLR)}.

\bibitem[{Hermann et~al.(2015)Hermann, Kočiský, Grefenstette, Espeholt, Kay, Suleyman, and Blunsom}]{hermann2015teaching}
Karl~Moritz Hermann, Tomáš Kočiský, Edward Grefenstette, Lasse Espeholt, Will Kay, Mustafa Suleyman, and Phil Blunsom. 2015.
\newblock Teaching machines to read and comprehend.
\newblock \emph{Advances in neural information processing systems}, 28.

\bibitem[{Horvitz et~al.(2024)Horvitz, Patel, Singh, Callison-Burch, McKeown, and Yu}]{horvitz-etal-2024-tinystyler}
Zachary Horvitz, Ajay Patel, Kanishk Singh, Chris Callison-Burch, Kathleen McKeown, and Zhou Yu. 2024.
\newblock \href {https://doi.org/10.18653/v1/2024.findings-emnlp.781} {{T}iny{S}tyler: Efficient few-shot text style transfer with authorship embeddings}.
\newblock In \emph{Findings of the Association for Computational Linguistics: EMNLP 2024}, pages 13376--13390, Miami, Florida, USA. Association for Computational Linguistics.

\bibitem[{Hu et~al.(2022)Hu, Shen, Wallis, Allen-Zhu, Li, Wang, Wang, and Chen}]{hu2022lora}
Edward~J Hu, Yelong Shen, Phillip Wallis, Zeyuan Allen-Zhu, Yuanzhi Li, Shean Wang, Lu~Wang, and Weizhu Chen. 2022.
\newblock \href {https://openreview.net/forum?id=nZeVKeeFYf9} {Lo{RA}: Low-rank adaptation of large language models}.
\newblock In \emph{International Conference on Learning Representations}.

\bibitem[{Izmailov et~al.(2018)Izmailov, Podoprikhin, Garipov, Vetrov, and Wilson}]{izmailov2018averaging}
Pavel Izmailov, Dmitrii Podoprikhin, Timur Garipov, Dmitry~P. Vetrov, and Andrew~Gordon Wilson. 2018.
\newblock \href {http://dblp.uni-trier.de/db/conf/uai/uai2018.html#IzmailovPGVW18} {Averaging weights leads to wider optima and better generalization}.
\newblock In \emph{Proceedings of the Thirty-Fourth Conference on Uncertainty in Artificial Intelligence (UAI)}, pages 876--885. AUAI Press.

\bibitem[{Khan et~al.(2021)Khan, Fleming, Schofield, Bishop, and Andrews}]{khan-etal-2021-deep}
Aleem Khan, Elizabeth Fleming, Noah Schofield, Marcus Bishop, and Nicholas Andrews. 2021.
\newblock \href {https://doi.org/10.18653/v1/2021.naacl-main.415} {A deep metric learning approach to account linking}.
\newblock In \emph{Proceedings of the 2021 Conference of the North American Chapter of the Association for Computational Linguistics: Human Language Technologies}, pages 5275--5287, Online. Association for Computational Linguistics.

\bibitem[{Klimt and Yang(2004)}]{klimt2004enron}
Bryan Klimt and Yiming Yang. 2004.
\newblock The enron corpus: A new dataset for email classification research.
\newblock In \emph{European Conference on Machine Learning}, pages 217--226. Springer.

\bibitem[{Li and Liang(2021)}]{li-liang-2021-prefix}
Xiang~Lisa Li and Percy Liang. 2021.
\newblock \href {https://doi.org/10.18653/v1/2021.acl-long.353} {Prefix-tuning: Optimizing continuous prompts for generation}.
\newblock In \emph{Proceedings of the 59th Annual Meeting of the Association for Computational Linguistics and the 11th International Joint Conference on Natural Language Processing (Volume 1: Long Papers)}, pages 4582--4597, Online. Association for Computational Linguistics.

\bibitem[{Liu et~al.(2024{\natexlab{a}})Liu, Lin, Hewitt, Paranjape, Bevilacqua, Petroni, and Liang}]{liu-etal-2024-lost}
Nelson~F. Liu, Kevin Lin, John Hewitt, Ashwin Paranjape, Michele Bevilacqua, Fabio Petroni, and Percy Liang. 2024{\natexlab{a}}.
\newblock \href {https://doi.org/10.1162/tacl_a_00638} {Lost in the middle: How language models use long contexts}.
\newblock \emph{Transactions of the Association for Computational Linguistics}, 12:157--173.

\bibitem[{Liu et~al.(2021)Liu, Ji, Fu, Tam, Du, Yang, and Tang}]{liu2021ptuning}
Xiao Liu, Kaixuan Ji, Yicheng Fu, Weng~Lam Tam, Zhengxiao Du, Zhilin Yang, and Jie Tang. 2021.
\newblock P-tuning: Prompt tuning can be comparable to fine-tuning.
\newblock \emph{arXiv preprint arXiv:2103.10385}.

\bibitem[{Liu et~al.(2024{\natexlab{b}})Liu, Diddee, and Ippolito}]{liu-etal-2024-customizing-large}
Xinyue Liu, Harshita Diddee, and Daphne Ippolito. 2024{\natexlab{b}}.
\newblock \href {https://aclanthology.org/2024.inlg-main.34} {Customizing large language model generation style using parameter-efficient finetuning}.
\newblock In \emph{Proceedings of the 17th International Natural Language Generation Conference}, pages 412--426, Tokyo, Japan. Association for Computational Linguistics.

\bibitem[{Ma et~al.(2023)Ma, Zhang, Bian, Liu, Zhang, Zhao, Zhang, Fu, Hu, and Wu}]{zhang2023fewshot}
Huan Ma, Changqing Zhang, Yatao Bian, Lemao Liu, Zhirui Zhang, Peilin Zhao, Shu Zhang, Huazhu Fu, Qinghua Hu, and Bingzhe Wu. 2023.
\newblock \href {http://arxiv.org/abs/2303.13217} {Fairness-guided few-shot prompting for large language models}.

\bibitem[{Matena and Raffel(2022)}]{matena2022merging}
Vrinda Matena and Colin Raffel. 2022.
\newblock \href {https://arxiv.org/abs/2209.14865} {Merging models with different initialization paths}.
\newblock \emph{arXiv preprint arXiv:2209.14865}.

\bibitem[{Min et~al.(2022)Min, Lewis, Hajishirzi, and Zettlemoyer}]{min2022rethinking}
Sewon Min, Mike Lewis, Hannaneh Hajishirzi, and Luke Zettlemoyer. 2022.
\newblock \href {https://arxiv.org/abs/2202.12837} {Rethinking the role of demonstrations: What makes in-context learning work?}
\newblock In \emph{Proceedings of the Conference on Empirical Methods in Natural Language Processing (EMNLP)}.

\bibitem[{Mukherjee and Dušek(2024)}]{mukherjee2024textstyletransferintroductory}
Sourabrata Mukherjee and Ondrej Dušek. 2024.
\newblock \href {http://arxiv.org/abs/2407.14822} {Text style transfer: An introductory overview}.

\bibitem[{OpenAI(2023)}]{openai2023gpt4}
OpenAI. 2023.
\newblock Gpt-4.
\newblock \url{https://platform.openai.com/docs/models/gpt-4}.
\newblock Accessed: YYYY-MM-DD.

\bibitem[{Ouyang et~al.(2022)Ouyang, Wu, Jiang, Almeida, Wainwright, Mishkin, Zhang, Agarwal, Slama, Ray, Schulman, Hilton, Kelton, Miller, Simens, Askell, Welinder, Christiano, Leike, and Lowe}]{NEURIPS2022_b1efde53}
Long Ouyang, Jeffrey Wu, Xu~Jiang, Diogo Almeida, Carroll Wainwright, Pamela Mishkin, Chong Zhang, Sandhini Agarwal, Katarina Slama, Alex Ray, John Schulman, Jacob Hilton, Fraser Kelton, Luke Miller, Maddie Simens, Amanda Askell, Peter Welinder, Paul~F Christiano, Jan Leike, and Ryan Lowe. 2022.
\newblock \href {https://proceedings.neurips.cc/paper_files/paper/2022/file/b1efde53be364a73914f58805a001731-Paper-Conference.pdf} {Training language models to follow instructions with human feedback}.
\newblock In \emph{Advances in Neural Information Processing Systems}, volume~35, pages 27730--27744. Curran Associates, Inc.

\bibitem[{Polo et~al.(2024)Polo, Weber, Choshen, Sun, Xu, and Yurochkin}]{polo2024tinybenchmarks}
Felipe~Maia Polo, Lucas Weber, Leshem Choshen, Yuekai Sun, Gongjun Xu, and Mikhail Yurochkin. 2024.
\newblock \href {http://arxiv.org/abs/2402.14992} {tinybenchmarks: evaluating llms with fewer examples}.

\bibitem[{Rao and Tetreault(2018)}]{rao-tetreault-2018-dear}
Sudha Rao and Joel Tetreault. 2018.
\newblock \href {https://doi.org/10.18653/v1/N18-1012} {Dear sir or madam, may {I} introduce the {GYAFC} dataset: Corpus, benchmarks and metrics for formality style transfer}.
\newblock In \emph{Proceedings of the 2018 Conference of the North {A}merican Chapter of the Association for Computational Linguistics: Human Language Technologies, Volume 1 (Long Papers)}, pages 129--140, New Orleans, Louisiana. Association for Computational Linguistics.

\bibitem[{Reif et~al.(2022)Reif, Ippolito, Yuan, Coenen, Callison-Burch, and Wei}]{reif-etal-2022-recipe}
Emily Reif, Daphne Ippolito, Ann Yuan, Andy Coenen, Chris Callison-Burch, and Jason Wei. 2022.
\newblock \href {https://doi.org/10.18653/v1/2022.acl-short.94} {A recipe for arbitrary text style transfer with large language models}.
\newblock In \emph{Proceedings of the 60th Annual Meeting of the Association for Computational Linguistics (Volume 2: Short Papers)}, pages 837--848, Dublin, Ireland. Association for Computational Linguistics.

\bibitem[{Rivera-Soto et~al.(2021)Rivera-Soto, Miano, Ordonez, Chen, Khan, Bishop, and Andrews}]{rivera-soto-etal-2021-learning}
Rafael~A. Rivera-Soto, Olivia~Elizabeth Miano, Juanita Ordonez, Barry~Y. Chen, Aleem Khan, Marcus Bishop, and Nicholas Andrews. 2021.
\newblock \href {https://doi.org/10.18653/v1/2021.emnlp-main.70} {Learning universal authorship representations}.
\newblock In \emph{Proceedings of the 2021 Conference on Empirical Methods in Natural Language Processing}, pages 913--919, Online and Punta Cana, Dominican Republic. Association for Computational Linguistics.

\bibitem[{Syed et~al.(2020)Syed, Verma, Srinivasan, Natarajan, and Varma}]{syed2020adapting}
Bakhtiyar Syed, Gaurav Verma, Balaji~Vasan Srinivasan, Anandhavelu Natarajan, and Vasudeva Varma. 2020.
\newblock Adapting language models for non-parallel author-stylized rewriting.
\newblock In \emph{Proceedings of the 34th AAAI Conference on Artificial Intelligence}.

\bibitem[{Touvron et~al.(2023)Touvron, Martin, Stone, Albert, Almahairi, Babaei, Bashlykov, Batra, Bhargava, Bhosale, Bikel, Blecher, Ferrer, Chen, Cucurull, Esiobu, Fernandes, Fu, Fu, Fuller, Gao, Goswami, Goyal, Hartshorn, Hosseini, Hou, Inan, Kardas, Kerkez, Khabsa, Kloumann, Korenev, Koura, Lachaux, Lavril, Lee, Liskovich, Lu, Mao, Martinet, Mihaylov, Mishra, Molybog, Nie, Poulton, Reizenstein, Rungta, Saladi, Schelten, Silva, Smith, Subramanian, Tan, Tang, Taylor, Williams, Kuan, Xu, Yan, Zarov, Zhang, Fan, Kambadur, Narang, Rodriguez, Stojnic, Edunov, and Scialom}]{touvron2023llama}
Hugo Touvron, Louis Martin, Kevin Stone, Peter Albert, Amjad Almahairi, Yasmine Babaei, Nikolay Bashlykov, Soumya Batra, Prajjwal Bhargava, Shruti Bhosale, Dan Bikel, Lukas Blecher, Cristian~Canton Ferrer, Moya Chen, Guillem Cucurull, David Esiobu, Jude Fernandes, Jeremy Fu, Wenyin Fu, Brian Fuller, Cynthia Gao, Vedanuj Goswami, Naman Goyal, Anthony Hartshorn, Saghar Hosseini, Rui Hou, Hakan Inan, Marcin Kardas, Viktor Kerkez, Madian Khabsa, Isabel Kloumann, Artem Korenev, Punit~Singh Koura, Marie-Anne Lachaux, Thibaut Lavril, Jenya Lee, Diana Liskovich, Yinghai Lu, Yuning Mao, Xavier Martinet, Todor Mihaylov, Pushkar Mishra, Igor Molybog, Yixin Nie, Andrew Poulton, Jeremy Reizenstein, Rashi Rungta, Kalyan Saladi, Alan Schelten, Ruan Silva, Eric~Michael Smith, Ranjan Subramanian, Xiaoqing~Ellen Tan, Binh Tang, Ross Taylor, Adina Williams, Jian~Xiang Kuan, Puxin Xu, Zheng Yan, Iliyan Zarov, Yuchen Zhang, Angela Fan, Melanie Kambadur, Sharan Narang, Aurelien Rodriguez, Robert Stojnic, Sergey Edunov, and Thomas
  Scialom. 2023.
\newblock \href {http://arxiv.org/abs/2307.09288} {Llama 2: Open foundation and fine-tuned chat models}.

\bibitem[{Wang et~al.(2023)Wang, Kordi, Mishra, Liu, Smith, Khashabi, and Hajishirzi}]{wang-etal-2023-self-instruct}
Yizhong Wang, Yeganeh Kordi, Swaroop Mishra, Alisa Liu, Noah~A. Smith, Daniel Khashabi, and Hannaneh Hajishirzi. 2023.
\newblock \href {https://doi.org/10.18653/v1/2023.acl-long.754} {Self-instruct: Aligning language models with self-generated instructions}.
\newblock In \emph{Proceedings of the 61st Annual Meeting of the Association for Computational Linguistics (Volume 1: Long Papers)}, pages 13484--13508, Toronto, Canada. Association for Computational Linguistics.

\bibitem[{Wortsman et~al.(2022)Wortsman, Ilharco, Gadre, Roelofs, Gontijo-Lopes, Morcos, Namkoong, Farhadi, Carmon, Kornblith, and Schmidt}]{pmlr-v162-wortsman22a}
Mitchell Wortsman, Gabriel Ilharco, Samir~Ya Gadre, Rebecca Roelofs, Raphael Gontijo-Lopes, Ari~S Morcos, Hongseok Namkoong, Ali Farhadi, Yair Carmon, Simon Kornblith, and Ludwig Schmidt. 2022.
\newblock \href {https://proceedings.mlr.press/v162/wortsman22a.html} {Model soups: averaging weights of multiple fine-tuned models improves accuracy without increasing inference time}.
\newblock In \emph{Proceedings of the 39th International Conference on Machine Learning}, volume 162 of \emph{Proceedings of Machine Learning Research}, pages 23965--23998. PMLR.

\bibitem[{Zhang et~al.(2024)Zhang, Cai, Li, Wu, Hou, and Abdul-Mageed}]{zhang-etal-2024-distilling}
Chiyu Zhang, Honglong Cai, Yuezhang Li, Yuexin Wu, Le~Hou, and Muhammad Abdul-Mageed. 2024.
\newblock \href {https://doi.org/10.18653/v1/2024.naacl-srw.21} {Distilling text style transfer with self-explanation from {LLM}s}.
\newblock In \emph{Proceedings of the 2024 Conference of the North American Chapter of the Association for Computational Linguistics: Human Language Technologies (Volume 4: Student Research Workshop)}, pages 200--211, Mexico City, Mexico. Association for Computational Linguistics.

\bibitem[{Zhou et~al.(2023)Zhou, Lu, Mishra, Brahma, Basu, Luan, Zhou, and Hou}]{zhou2023instruction}
Jeffrey Zhou, Tianjian Lu, Swaroop Mishra, Siddhartha Brahma, Sujoy Basu, Yi~Luan, Denny Zhou, and Le~Hou. 2023.
\newblock Instruction-following evaluation for large language models.
\newblock \emph{arXiv preprint arXiv:2311.07911}.

\end{thebibliography}

\clearpage
\appendix
\onecolumn
\noindent
\large{\textbf{Appendix}}
\section{Additional IFEval Results}
\label{sec:instruction-following-add}

\begin{table*}[htbp] 
\centering
\resizebox{\textwidth}{!}{%
\begin{tabular}{l|cccc|cccc|cccc}
\toprule
 & \multicolumn{4}{c|}{\textbf{LLaMa-3.1 8B}} 
 & \multicolumn{4}{c|}{\textbf{Mistral 7B}} 
 & \multicolumn{4}{c}{\textbf{Qwen-2.5 7B}} \\
\midrule
\textbf{Method} 
  & \textbf{Tana} & \textbf{Sally} & \textbf{Red. B} & \textbf{Red. C}
  & \textbf{Tana} & \textbf{Sally} & \textbf{Red. B} & \textbf{Red. C}
  & \textbf{Tana} & \textbf{Sally} & \textbf{Red. B} & \textbf{Red. C} \\
\midrule
Instruct Model (No FT)
  & \multicolumn{4}{c|}{{77.10}}
  & \multicolumn{4}{c|}{{57.67}}
  & \multicolumn{4}{c}{{70.55}}\\ 
\midrule
Prompting
  & 57.79 & 60.43 & 65.68 & 67.23
  & 42.16 & 42.40 & 33.14 & 39.11
  & 50.60 & 53.00 & 55.60 & 57.97 \\
Prompting (Long Context)
  & 55.82 & 58.18 & 64.61 & 62.79
  & 41.97 & 40.12 & 33.92 & 39.42
  & 49.80 & 50.00 & 54.45 & 55.50 \\
Direct LoRA FT
  & 54.10 & 54.44 & 62.67 & 69.14
  & 46.76 & 45.44 & 42.71 & 41.97
  & 50.20 & 50.00 & 58.13 & 60.03 \\
Model Soup (2:1)
  & 69.83 & 68.16 & 71.82 & 71.97
  & 52.92 & 49.12 & 46.47 & 43.11
  & 64.00 & 62.00 & 62.93 & 61.97 \\
\modelname
  & 71.82 & 71.74 & 73.22 & 71.94
  & 52.28 & 50.12 & 46.08 & 45.80
  & 65.00 & 64.00 & 62.97 & 63.23 \\
\bottomrule
\end{tabular}%
}
\caption{IFEval (Strict) Accuracy. Red. B = Reddit User B, Red. C = Reddit User C.}
\label{tab:instruction-following-add}
\end{table*}

\section{MMLU Results}
\label{sec:tinymmlu}

\begin{table*}[htbp] 
\centering
\small
\begin{tabular}{l|c|c|c}
\toprule
 & \multicolumn{1}{c|}{\textbf{LLaMa-3.1 8B}} 
 & \multicolumn{1}{c|}{\textbf{Mistral 7B}} 
 & \multicolumn{1}{c}{\textbf{Qwen-2.5 7B}} \\
\midrule
Instruct Model (No FT)
  & 63 & 55 & 60\\ 
\midrule
\modelname
  & 63 & 55 & 60 \\
\bottomrule
\end{tabular}%
\caption{Accuracy on tinyMMLU benchmark (5 shot).}
\label{tab:tinymmlu}
\end{table*}

\section{Authorship Attribution Classifier Accuracy}
\label{sec:class_acc}

\begin{table}[h]
  \centering
  \small
  {
    \begin{tabular}{l|r}
      \toprule
      \textbf{Persona Classifier} & \textbf{F1 score}\\ 
      \midrule
      Jeff & 0.967 \\
        Tana & 0.967 \\
        Sally & 0.987 \\
        BBC & 0.964 \\
        CNN & 0.964 \\ 
      \bottomrule
    \end{tabular}
  }
\caption{Authorship attribution classifier F1 score for each persona.}
\label{tab:classifier_acc}
\end{table}

\section{Additional Brand Adherence Results}
\label{sec:style-add}
\begin{table}[H]
\centering
\resizebox{\textwidth}{!}{%
    \begin{tabular}{l|cccc|cccc|cccc}
\toprule
 & \multicolumn{4}{c|}{\textbf{LLaMa-3.1 8B}} 
 & \multicolumn{4}{c|}{\textbf{Mistral 7B}} 
 & \multicolumn{4}{c}{\textbf{Qwen-2.5 7B}} \\
\midrule
        \textbf{Method} & \textbf{Tana} & \textbf{Sally} & \textbf{Red. B} & \textbf{Red. C} & \textbf{Tana} & \textbf{Sally} & \textbf{Red. B} & \textbf{Red. C} & \textbf{Tana} & \textbf{Sally} & \textbf{Red. B} & \textbf{Red. C} \\
\midrule
        Prompting & 0.90 & 0.97 & 0.95 & 1.00 & 0.83 & 0.79 & 0.89 & 1.00 & 1.00 & 0.87 & 0.95 & 1.00 \\
        Prompting (Long Context) & 0.90 & 0.97 & 0.96 & 1.00 & 0.83 & 0.89 & 1.00 & 1.00 & 0.93 & 0.95 & 0.97 & 1.00 \\
        Direct LoRA Fine-tuning & 0.88 & 0.99 & 0.91 & 1.00 & 0.79 & 0.84 & 0.92 & 0.90 & 0.84 & 0.95 & 0.91 & 0.97 \\
        Model Soup (2:1) & 0.87 & 0.98 & 0.90 & 1.00 & 0.81 & 0.74 & 0.84 & 0.94 & 0.96 & 0.94 & 0.91 & 0.99 \\
        BrandAdaptedLM & 0.94 & 0.99 & 0.92 & 1.00 & 0.75 & 0.85 & 0.90 & 0.90 & 0.95 & 0.93 & 0.97 & 0.99 \\ 
\midrule
    \end{tabular}
}
    \caption{F1 Score of Brand Adherence Based on Authorship Attribution.}
    \label{tab:style_transfer_add}
\end{table}

\end{document}